\documentstyle[fullname,lingmacros]{article}

\title{Restricting the Weak-Generative Capacity of Synchronous
Tree-Adjoining Grammars}

\author{Stuart M. Shieber\\
	Division of Applied Sciences\\
	Harvard University\\
	33 Oxford Street\\
	Cambridge, MA \quad 02138 \\
	{\tt shieber@das.harvard.edu}}

\date{August 30, 1994}

\input{psfig-scale}

\newcommand{\eqpunc}[1]{{\makebox[0pt][l]{\qquad\rm{#1}}}}

\newcommand{\seq}[1]{{\langle{#1}\rangle}}
\newcommand{\set}[1]{{\{{#1}\}}}
\newcommand{\link}{\frown}
\newcommand{\lnk}[1]{\raisebox{.5ex}{\tiny\fbox{#1}}}

\newcommand{\DD}{{\cal D}}
\newcommand{\NP}{\mbox{\it NP}}

\newcommand{\used}[1]{{`#1'}}
\newcommand{\word}[1]{{`#1'}}

\newcommand{\qbox}[1]{\begin{center}\mbox{#1}\end{center}}

\makeatletter
\def\thebibliography#1{\subsubsection*{References\@mkboth
 {References}{References}}\list
 {}{\setlength{\labelwidth}{0pt}\setlength{\leftmargin}{\parindent}
 \setlength{\itemindent}{-\parindent}}
 \def\newblock{\hskip .11em plus .33em minus -.07em}
 \sloppy\clubpenalty4000\widowpenalty4000
 \sfcode`\.=1000\relax}
\makeatother

\begin{document}

\begin{titlepage}

\maketitle
\thispagestyle{empty}

\begin{abstract}
The formalism of synchronous tree-adjoining grammars, a variant of
standard tree-adjoining grammars (TAG), was intended to allow the use
of TAGs for language transduction in addition to language
specification.  In previous work, the definition of the transduction
relation defined by a synchronous TAG was given by appeal to an
iterative rewriting process.  The rewriting definition of derivation
is problematic in that it greatly extends the expressivity of the
formalism and makes the design of parsing algorithms difficult if not
impossible.

We introduce a simple, natural definition of synchronous
tree-adjoining derivation, based on isomorphisms between standard
tree-adjoining derivations, that avoids the expressivity and
implementability problems of the original rewriting definition.  The
decrease in expressivity, which would otherwise make the method
unusable, is offset by the incorporation of an alternative definition
of standard tree-adjoining derivation, previously proposed for
completely separate reasons, thereby making it practical to entertain
using the natural definition of synchronous derivation.  Nonetheless,
some remaining problematic cases call for yet more flexibility in the
definition; the isomorphism requirement may have to be relaxed.  It
remains for future research to tune the exact requirements on the
allowable mappings.
\end{abstract}

\begin{verse}\small
\end{verse}

\vfill

{\noindent\small This paper is to appear in {\em Computational
Intelligence}, and is available through the Computation and Language
e-print archive as cmp-lg/9404003.}

\end{titlepage}

\tableofcontents

\newpage

\section{Introduction}

The formalism of synchronous tree-adjoining grammars \cite{synch-tag},
a variant of standard tree-adjoining grammars (TAG), was intended to
allow the use of TAGs for language transduction in addition to
language specification.  Synchronous TAGs specify relations between
language pairs; each language is specified with a standard TAG, and
the pairing between strings in the separate languages is specified by
synchronizing the two TAGs through linking pairs of elementary trees.

This paper concerns the formal definitions underlying synchronous
tree-adjoining grammars.  In previous work \cite{synch-tag}, the
definition of the transduction relation defined by a synchronous TAG
was given by appeal to an iterative rewriting process, much like the
iterative rewriting of sentential forms defined by a context-free
grammar except that the syntactic objects generated by the rewriting
process were derived trees rather than strings.  This sort of
rewriting definition of derivation is problematic for several reasons.
First, the weak-generative expressivity of TAGs is increased through
the synchronization in the sense that the projection of the string
pairs onto a single component, although the strings in that component
are specified with a TAG, may not form a tree-adjoining language
(TAL).  Second, the lack of a simple recursive characterization of the
derivation --- a role filled by derivation trees for standard TAGs ---
makes the design of parsing algorithms difficult if not impossible.

In this paper, we describe how synchronous TAG derivation can be
redefined so as to eliminate these problems.  The redefinition relies
on an independent redefinition of the notion of tree-adjoining
derivation \cite{full-alt-deriv} that was motivated
completely independently of worries about the generative capacity of
synchronous TAGs, but which happens to solve this problem in an
elegant manner.  Furthermore, it allows for existing parsing
algorithms to be generalized to synchronous TAG transduction in a
natural way.  However, because certain derivations in the rewriting
sense have no analogue under the new definition, some linguistic
analyses may no longer be statable.  We comment on some possible
negative ramifications of this fact.

\section{The Rewriting Definition of Derivation}

The original definition of derivation for synchronous TAGs was based
on the iterative rewriting of one derived tree pair into another.  In
this section, we provide a more precise description of this approach
to derivation, along with a discussion of its problems.  First,
however, we digress to discuss some notational issues.

\subsection{Notation}

We assume in this and later sections a general familiarity with
tree-adjoining grammars and their formal foundations, as described,
for instance, by
\namecite{v87}.

We will use the following notational conventions for synchronous TAGs
and related notions.  A synchronous TAG $G$ will be given as a set of
triples $\set{\seq{L_i, R_i, \link_i}}$ where the $L_i$ and $R_i$ are
elementary trees, both initial and auxiliary, forming two component
TAGs $G_L = \set{L_i}$ and $G_R = \set{R_i}$, and $\link_i$ is the
linking relation between tree addresses in $L_i$ and $R_i$.  Such a
grammar is intended to define a language of pairs $L(G) =
\set{\seq{l_i, r_i}}$; the exact manner in which $L(G)$ is determined
is the subject of this paper.  We will use the notation $x_L$ and
$x_R$ to notate the projection of a pair $x$ onto its left and right
components, respectively, and generalize this notation to the first
and second components of a triple and pointwise on sets of pairs and
triples.  Thus, the notations $G_L$ and $G_R$ previously introduced
for the left and right component grammars are consistent with this
notation.

\subsection{The Rewriting Process}

The rewriting process proceeds by choosing an initial tree pair
$\seq{I_L, I_R, \link}$ to be the {\it current derived tree pair} and
repeatedly performing the following steps:

\begin{enumerate}

\item	Choose a link ${t_L \link t_R}$ between two nodes in the
current derived tree pair.

\item	Choose an auxiliary tree pair $\seq{A_L, A_R, \link'}$ from
the grammar such that $A_L$ can adjoin at $t_L$ in $I_L$ and $A_R$ can
adjoin at $t_R$ in $I_R$.

\item	Modify the current derived tree pair by adjoining the chosen
trees at the end of the chosen link, yielding the modified derived
tree pair $$\seq{I_L[A_L/t_L], I_R[A_R/t_R], \link''}\eqpunc{.}$$ This
becomes the new current derived tree pair.

\end{enumerate}

The operation $I[A/t]$ used above takes a tree $I$, an auxiliary tree
$A$, and an address $t$ in $I$ and yields the result of adjoining $A$
at address $t$ in $I$.  (The generalization to allow for substitution
as well as adjunction as a primitive operation --- both in this
notation and the definition of derivation --- should be clear.)  A
formal definition for this operation is given by Vijay-Shanker
\shortcite[page 15]{v87} and by Shieber and Schabes
\shortcite[appendix]{full-alt-deriv}.

The definition of the link relation in the derived tree pair $\link''$
is as follows: All links in $\link$ and $\link'$ are included in
$\link''$ (after suitable readdressing) except that the chosen link in
$\link$ is not itself included in $\link''$.  Other links that impinge
on the nodes at the end of the chosen link are retained in the derived
tree pair; they link to the root or foot of the newly adjoined tree as
determined by whether the link itself is viewed as impinging on the
top or the bottom of the node.\footnote{In previous work, links were
typically thought of as impinging on the top of a node unless
otherwise stated.  We will retain that convention here.  Further
flexibility can be obtained by allowing each link to specify whether
it links to the top or bottom of the nodes.  Thus, the link relation
$\link_i$ in a triple $\seq{L_i, R_i, \link_i}$ can be thought of as
being of type $dom(L_i) \times \set{ \uparrow, \downarrow} \times
dom(R_i) \times \set{\uparrow, \downarrow}$, where $dom(A)$ is the set
of tree addresses in the tree $A$ and $\uparrow$ and $\downarrow$
serve as markers to specify whether the link impinges on the top or
bottom, respectively, of the specified node.  All of this machinery
becomes superfluous, however, in the context of the natural definition
of derivation given in Section~\ref{sec:natural}.}

\subsection{An Example of Rewriting}

By way of example, we present a sample synchronous TAG that transduces
between a tiny fragment of English and a corresponding ``logical
form'' semantic representation.

\begin{figure}
\qbox{\psfig{figure=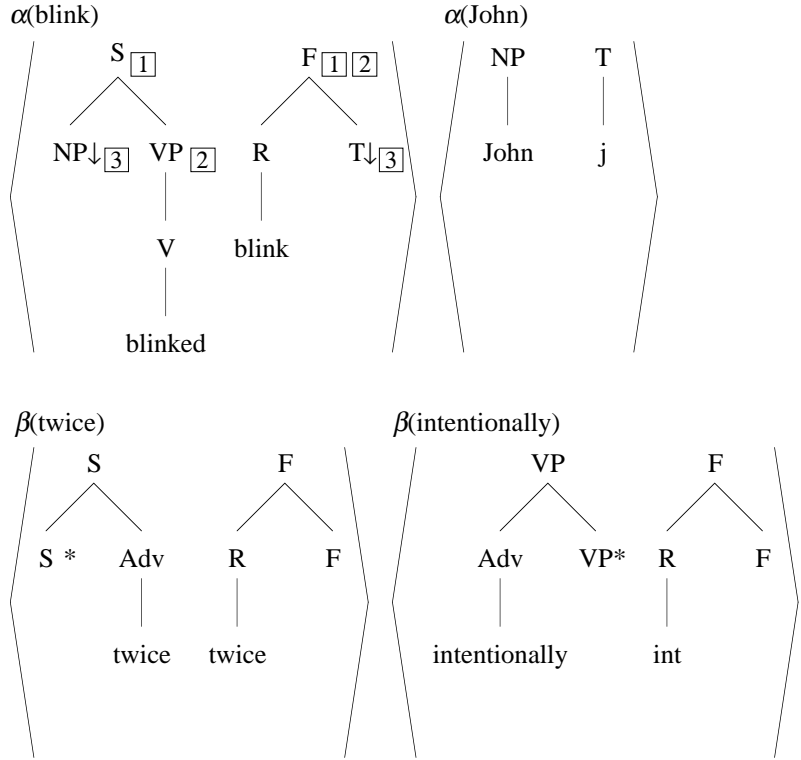}}
\caption{A synchronous TAG that describes the semantic ambiguity of the
sentence \used{John intentionally blinked twice}.}
\label{fig:blink-gram}
\end{figure}

Figure~\ref{fig:blink-gram} shows the sample synchronous TAG composed
of a set of tree pairs, each with a left element that is part of an
English TAG fragment and a right element that is part of a TAG
fragment for the logical form language.  Thus, the tree pair labeled
$\alpha(John)$ pairs a noun phrase ($\NP$) initial tree dominating the
proper noun \word{John} with a logical term ($T$) dominating the constant
$john$.  Similarly, the tree pair $\alpha(blink)$ pairs a verb tree
for \word{blinked} with a tree for a formula ($F$) constructed as the
predication of the relation ($R$) given by the symbol $blink$ to an
unspecified argument term.

Rather than present the elements of the grammar as triples, we notate
the links between nodes with diacritics.  Thus, the $\alpha(blink)$
tree pair implicitly incorporates the link relation between tree
addresses given by
\[ \begin{array}{rcl}
	\epsilon \link \epsilon \\
	2 \link \epsilon \\
	1 \link 2
   \end{array}
\]
These three links are marked with the diacritics \lnk{1}, \lnk{2}, and
\lnk{3} respectively.

\begin{figure}
\qbox{\psfig{figure=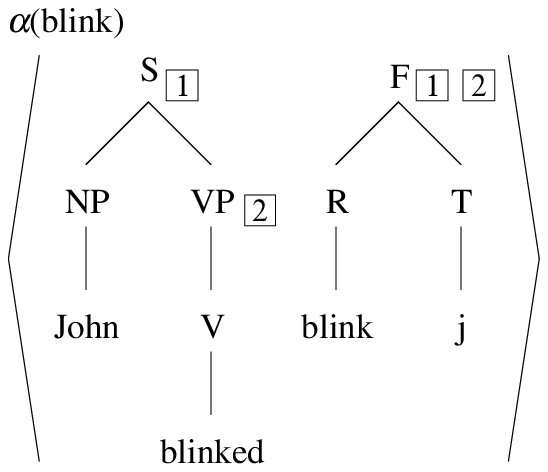,clip=}}
\caption{A tree pair derived by operation of the $\alpha(John)$ tree
at link \protect\lnk{3} in the $\alpha(blink)$ tree.  The pair specifies the
transduction between the string \used{John blinks} and its logical form
$blink(john)$.}
\label{fig:john-blinks}
\end{figure}

The \lnk{3} link, for instance, connects the $\NP$ node in the left
tree at address 1 with the $T$ node in the right at address 2, thereby
allowing the two trees of another tree pair to operate respectively at
these two nodes.  Since the two nodes are substitution nodes (as
conventionally marked by the $\downarrow$), the operations on this
link would be substitutions at both ends.  For example, the initial
tree pair $\alpha(John)$ can operate at this link, yielding the tree
pair given in Figure~\ref{fig:john-blinks}.  Note that the remaining
links in the $\alpha(blink)$ tree labeled
\lnk{1} and \lnk{2} are preserved in the derived tree pair.

\begin{figure}
\qbox{\psfig{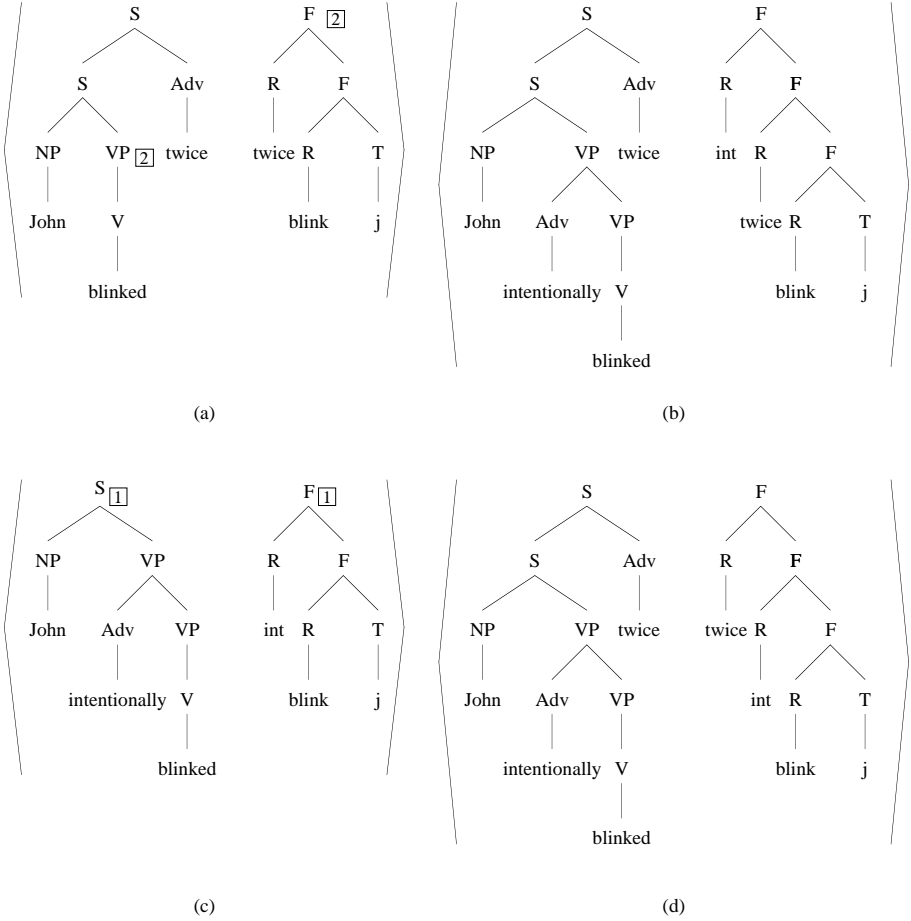}}
\caption{Derived tree pairs from the grammar of
Figure~\protect\ref{fig:blink-gram}.  The derivation of the meaning
$int(twice(blink(j)))$ proceeds through the derived tree pairs in (a)
and (b).  The derivation of the meaning $twice(int(blink(j)))$
proceeds through the derived tree pairs in (c) and (d).}
\label{fig:blink-derivs}
\end{figure}

Continuing on in this way, the resultant derived tree pair can be
further acted upon, say, by the base pair $\beta(twice)$, whose
elements can adjoin at the ends of the \lnk{1} link, yielding the
derived tree pair in Figure~\ref{fig:blink-derivs}a.  The issue of how
to handle multiple links impinging on the same node becomes relevant
here, since the right end of the remaining link \lnk{2} in the derived
tree pair impinges on a node at which adjunction has just occurred.
Should the link now impinge on the root or the foot node of the tree
adjoined at that node?  We place the link at the root, as stipulated
above, so that further rewriting of the \lnk{2} link, say with the
adverbial tree pair $\beta(intentionally)$ leads to the derived tree
pair in Figure~\ref{fig:blink-derivs}b, corresponding to the string
\used{John intentionally blinked twice}.  In the associated logical form,
the predication of $int$ has scope over the proposition
$twice(blink(john))$, and the sentence is taken to describe a single
intentional act of blinking twice.  Had the two links been rewritten
in the other order --- link \lnk{2} first, yielding the pair in
Figure~\ref{fig:blink-derivs}c, and then link \lnk{1} yielding the
pair in Figure~\ref{fig:blink-derivs}d --- the generated logical form
$twice(int(blink(j)))$ describes two intentional acts each of single
blinkings.

Thus, this grammar manifests the ambiguity in the sentence \used{John
intentionally blinked twice}. Note that the ambiguity arises from the
ability to perform two rewriting steps at the same node, the root $F$
node in the logical form tree $\alpha(blink)_R$ corresponding to the
word \word{blinked}.

\subsection{Problems with the Rewriting Definition}
\label{sec:problems}

There are two problems with the rewriting definition of synchronous
TAGs, having to do with the expressivity and implementability of the
formalism under that definition.

\subsubsection{Expressivity}

Synchronous TAGs under this definition may specify
non-tree-adjoining languages.  More precisely stated, given a grammar
$G$, although, by definition, $L(G_L)$ is a tree-adjoining language,
$L(G)_L$ may not be.

\begin{figure}
\qbox{\psfig{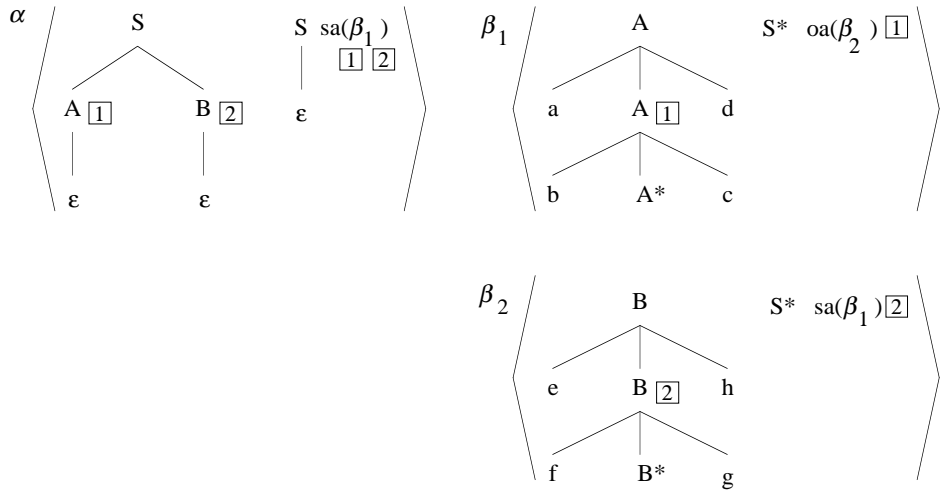}}
\caption{A synchronous TAG for a non-tree-adjoining language
$a^nb^nc^nd^ne^nf^ng^nh^n$.}
\label{fig:bad-gram}
\end{figure}

\begin{figure}
\qbox{\psfig{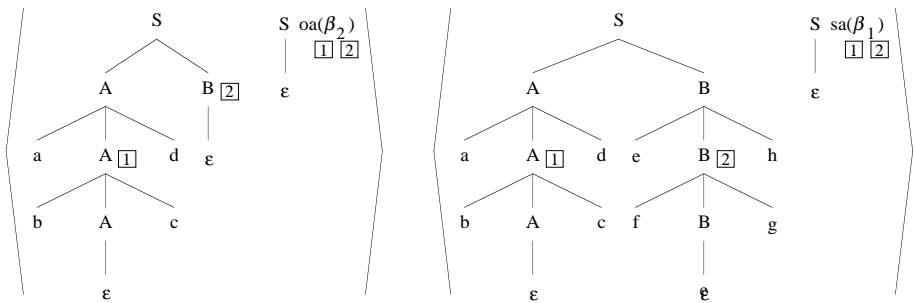}}
\caption{Steps in the derivation of {\it abcdefgh}.  The left
derived tree pair has a remaining obligatory adjoining constraint,
which when satisfied yields the right derived tree pair.}
\label{fig:eight-deriv}
\end{figure}

\begin{figure}
\qbox{\psfig{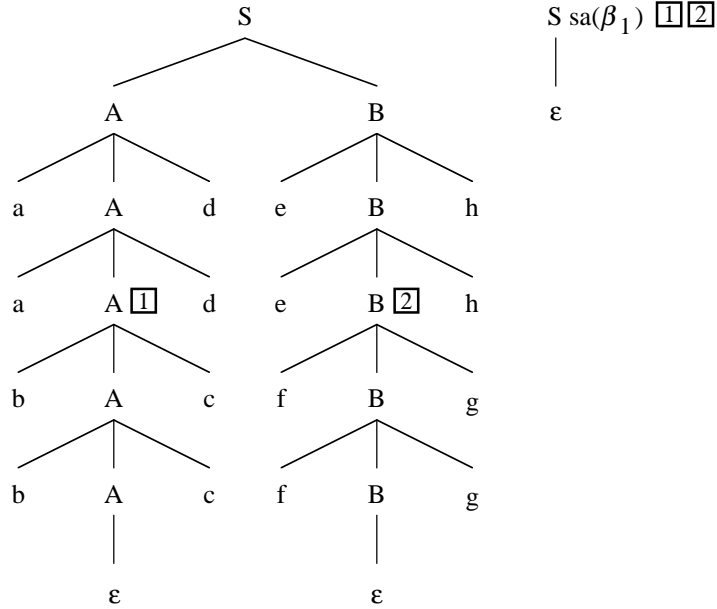}}
\caption{Derived tree pair for $a^2b^2c^2d^2e^2f^2g^2h^2$.}
\label{fig:bad-deriv}
\end{figure}

A simple example of a synchronous TAG that can be projected onto a
non-TAL is given in Figure~\ref{fig:bad-gram}.  This grammar specifies
the string relation that pairs all strings of the form
$a^nb^nc^nd^ne^nf^ng^nh^n$ with the empty string.  Its projection onto
its first component is, therefore, a non-tree-adjoining language.
Figure~\ref{fig:eight-deriv} shows the steps in the derivation of the
$n=1$ case.  The derived tree pair for the $n=2$ case is given in
Figure~\ref{fig:bad-deriv}.

\subsubsection{Implementability}

In addition to the expressivity problem, there is no natural way to
use a synchronous grammar for transduction under this definition.  To
use a synchronous TAG $G$ for transduction, a given string $w_L$ is to
be transduced to $w_R$ just in case $\seq{w_L, w_R} \in L(G)$.  This
requires, intuitively speaking, parsing of the string $w_L$ relative
to $G_L$ yielding a derivation $D_L$, reconstruction of the
synchronous (rewriting) derivation $D_S$ from $D_L$, and finally,
generation of the string $w_R$ according to this reconstructed
derivation.  Schematically, the process can be depicted as proceeding
thus:
\[w_L \longrightarrow D_L \longrightarrow D_S \longrightarrow w_R\]
Unfortunately, the structure of a synchronous derivation bears no
uniform relationship to the kind of derivation postulated for standard
TAGs.  (This point is discussed further in the next section.)  Thus,
if a standard TAG parsing algorithm is used for the first step in the
process (so that $D_L$ is a traditional TAG derivation tree), the
second step is not well defined.  It is therefore not clear how
synchronous TAGs can be effectively used under this definition of
derivation.

Note that this point is independent of whether the three conceptual
phases of processing are interleaved in time.  The possibility to
interleave the computations of the phases does not make their
definition any simpler.

\section{The Natural Definition of Derivation}
\label{sec:natural}

The notion of derivation just presented for synchronous TAGs is quite
nonstandard for the TAG literature in being ``flat'' and rewriting
oriented.  Recall that the standard definition of TAG derivation, due
to \namecite{v87}, is hierarchically structured in terms of {\it
derivation trees}, trees that serve to characterize the operations
required to construct a particular derived tree, and hence its yield.

TAG derivation trees are composed of nodes, conventionally notated as
$\eta$, possibly in its subscripted variants.  The parent of a node
$\eta$ in a derivation tree will be written $parent(\eta)$, and the
tree that the node marks adjunction of will be notated $tree(\eta)$.
The tree $tree(\eta)$ is to be adjoined into its parent
$tree(parent(\eta))$ at an address specified on the arc in the tree
linking the two; this address is notated $addr(\eta)$.  (Of course the
root node has no parent or address; the $parent$ and $addr$ functions
are partial.)

A derivation tree is well-formed if for each arc in the derivation
tree from $\eta$ to $parent(\eta)$ labeled with $addr(\eta)$, the tree
$tree(\eta)$ is an auxiliary tree that can be adjoined at the node
$addr(\eta)$ in $tree(parent(\eta))$.  (Alternatively, $tree(\eta)$ is
an initial tree that can be substituted at the node $addr(\eta)$ in
$tree(parent(\eta))$.)  Furthermore, and without loss of expressivity,
it is standard to exclude multiple sibling arcs specifying operations
at the same tree address in the same tree.  This exclusion makes the
definition of the derived tree for a given derivation tree
determinate.

A derivation tree specifies a derived tree by virtue of the normal
definitions for adjunction and substitution.  The language of a TAG
$G$ is then the set of strings that are the yields of derived trees
specified by derivation trees that are well-formed according to $G$.
We define the function $\DD$ from derivation trees to the derived
trees they specify, according to the following recursive definition:

\[
\DD(D) = \left\{
\begin{array}{l}
tree(\eta) \\
\mbox{\qquad if $D$ is a trivial tree of one node $\eta$}\\[1ex]
tree(\eta)[\DD(D_1)/t_1, \DD(D_2)/t_2, \ldots, \DD(D_k)/t_k] \\
\mbox{\qquad if\begin{tabular}[t]{l}
		$D$ is a tree with root node $\eta$ \\
		and with $k$ child subtrees $D_1, \ldots, D_k$
		\end{tabular}}
\end{array}
\right.
\]

Here $I[A_1/t_1, \ldots, A_k/t_k]$ specifies the simultaneous
adjunction (or substitution) of trees $A_1$ through $A_k$ at $t_1$
through $t_k$, respectively, in $I$.  Using the definitions of
Vijay-Shanker \shortcite{v87}, this is well defined only as long as
the $t_i$ are disjoint, hence the need for the aforementioned
exclusion.

A definition along these lines for synchronous TAGs would be quite
natural.  We would have derivation trees that specify at each node an
elementary tree {\it pair}, with arcs labeled by {\em pairs} of tree
addresses (such that the two addresses are linked in the parent
elementary tree pair).  A function from derivation trees to the
derived tree pairs they specify --- a generalization of the $\DD$
function defined above --- would then be used to generate the derived
trees and the language of a synchronous grammar.

It should be obvious that such a synchronous derivation tree can be
trivially restated as a pair of standard derivation trees, further
simplifying the definition of synchronous TAG derivation.  This leads
to the following definition of synchronous TAG derivation.  A
derivation is a pair $\seq{D_L, D_R}$ where

\begin{enumerate}

\item $D_L$ is a well-formed derivation tree relative to $G_L$.

\item $D_R$ is a well-formed derivation tree relative to $G_R$.

\item $D_L$ and $D_R$ are isomorphic.  That is, there is  a one-to-one
onto mapping $f$ from the nodes of $D_L$ to the nodes of $D_R$ that
preserves dominance, i.e., if $f(\eta_l) = \eta_r$ then
$f(parent(\eta_l)) = parent(\eta_r)$.

\item The isomorphic operations are sanctioned by links in tree pairs.
That is, if $f(\eta_l) = \eta_r$, then there is a tree pair
$\seq{tree(\eta_l), tree(\eta_r), \link'}$ in $G$.  Furthermore, if
$\eta_l$ has a parent, then there is a tree pair
$$\seq{tree(parent(\eta_l)), tree(parent(\eta_r)), \link}$$ in $G$ and
$addr(\eta_l) \link addr(\eta_r)$.

\end{enumerate}

\noindent This, then, is the most natural definition of synchronous
tree-adjoining derivation, as it is the natural generalization of the
definition of derivation for standard TAGs.  It merely requires that
there be two derivations that are separately well-formed and
appropriately synchronized as specified by the links.

Several aspects of this definition are noteworthy. First, the derived
tree pair for a derivation $\seq{D_L, D_R}$ is $\seq{\DD(D_L),
\DD(D_R)}$.  Second, the definition does not require extra linking
information specifying whether the link impinges on the top or bottom
of the linked nodes.  It is completely declarative; no vestiges remain
of the rewriting definition.  Finally, it solves the two problems of
expressivity and implementability mentioned above, as described in the
next section.

\section{Advantages of the Natural Definition}
\label{sec:advantages}

We show in this section that the natural definition of synchronous
derivation solves the two problems described in
Section~\ref{sec:problems}.

\subsection{Expressivity}

Under the revised definition of synchronous derivation, only
tree-adjoining languages can be expressed by a synchronous TAG.  To
see why, we look first at the problematic example of
Figure~\ref{fig:bad-gram}, and then turn to a general argument.

Under the new definition, adjoining constraints are no longer
inherited in an overall derived tree being generated incrementally in
the flat rewriting process.  Rather, they apply to the auxiliary trees
that directly adjoin to the node.  Thus, in the grammar of
Figure~\ref{fig:bad-gram}, the links in the auxiliary trees can never be
operated on.  For instance, the link in $\beta_1$ requires $\beta_2$
to be adjoined there, but its corresponding left half cannot adjoin at
the left end of the link.  Similarly, the link in $\beta_2$ is useless
as well.  Thus, the only well-formed derivation is the one with no
adjunctions whatsoever; the language of the grammar includes the
single string pair $\seq{\epsilon, \epsilon}$ generated by its initial
tree pair.

\begin{figure}
\qbox{\psfig{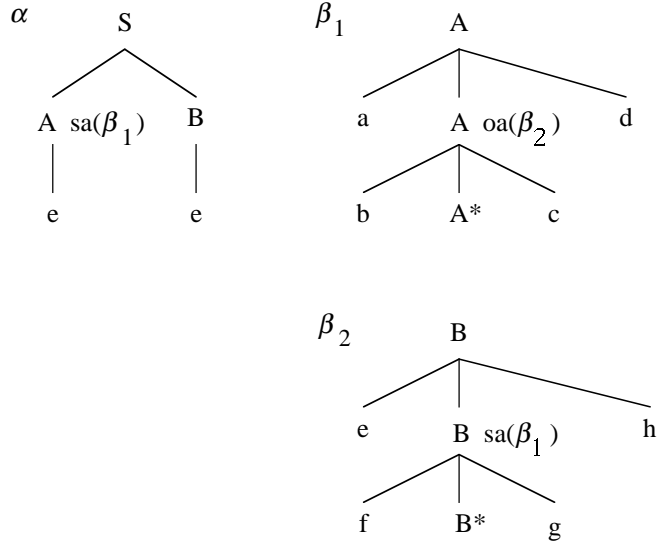}}
\caption{The left projection of the grammar of
Figure~\protect\ref{fig:bad-gram}.}
\label{fig:bad-project}
\end{figure}

In general under the revised definition, the left-projection language,
say, of a synchronous TAG is specifiable by a pure TAG by simply
mapping any adjoining constraints on the right trees to corresponding
ones on the linked nodes on the left and projecting the grammar on its
left component.  (The example of Figure~\ref{fig:bad-gram}, so
projected, is the normal TAG given in Figure~\ref{fig:bad-project},
which specifies the language containing only the empty string as
expected.)

Alternatively, the TAL nature of synchronous TAGs under this
definition can be easily shown by reduction to tree-set-local
multicomponent TAGs (MCTAG), which are known to generate only
tree-adjoining languages.\footnote{The observation that synchronous
TAGs under the new definition should be reducible to MCTAG was brought
to our attention by Owen Rambow.} Each elementary tree pair in the
synchronous TAG corresponds to an elementary tree set in the MCTAG.
To ensure that left-hand trees are not adjoined into right-hand trees
and vice versa, the node labels on the left- and right-hand trees are
uniformly renamed apart.  Each node in a left-hand tree is marked with
a selective adjoining constraint that allows adjunction only of
certain elementary tree sets.  For each link that impinges on the
node, and each tree pair that can operate on that link, the
corresponding tree set is allowed by the SA constraint.  Similar
constraints are added to each right-hand node.  Finally, for each pair
of nonterminals that root the trees in an initial tree pair, a new
elementary tree is constructed rooted in a new nonterminal symbol not
used elsewhere with two nonterminal children labeled by the left and
right root nonterminals of the initial tree pair and which are to be
filled by substitution.

Since any synchronous TAG can be reduced to a tree-set-local MCTAG,
the languages generated by synchronous TAGs are at most the
tree-adjoining languages.  The converse inclusion is obvious.

\subsection{Implementability}

Another advantage of the new definition of synchronous derivation is
in its utility for implementation of synchronous TAG transducers.
Recall that under the rewriting definition, the structure of a
synchronous derivation bears no uniform relationship to the kind of
derivation postulated for standard TAGs and therefore recovered by
standard TAG parsing algorithms.  Thus, the second step in the
schematic process
\[w_L \longrightarrow D_L \longrightarrow D_S \longrightarrow w_R\]
is not well defined.  Under the natural definition, however, the
synchronous derivation $D_S$ is just $\seq{D_L,D_R}$.  This close
relation between a synchronous TAG derivation and derivations for the
left and right projected grammars makes synchronous transduction
straightforward.  Any method for parsing that generates a standard
derivation tree for a grammar can be applied to parse a string $w_L$
relative to the left projection grammar.  The resultant derivation is
isomorphic to the derivation tree for the right projection grammar,
where the mapping is given directly by the synchronous grammar.  The
right projection derivation is thus easily constructed, and the
corresponding derivation tree can be computed directly.
Schematically, the process looks like this:
\[w_L \longrightarrow D_L \longrightarrow D_S (= \seq{D_L,D_R})
\longrightarrow D_R \longrightarrow w_R\]

This methodology applies even under the view of synchronous TAG
derivations to be described in Section~\ref{sec:multiple}.  For
instance, \namecite{full-alt-deriv} describe a parsing method for
standard TAGs that can be used to construct derivation trees on the
fly while parsing.  A simple modification of the method can construct
the isomorphic derivation tree for the object grammar of a
transduction.  In fact, this redefinition has allowed for the first
implementation of synchronous TAG processing, due to Onnig
Dombalagian.  This implementation was based on the inference-based TAG
parser that we have presented elsewhere \cite{full-alt-deriv}.

\section{Problems with the Natural Definition}

Along with the advantages of the new definition of synchronous TAG
derivation, new problems are introduced as well.  First, the exclusion
of multiple adjunctions at a single address is problematic for
synchronous TAG derivations.  Second, the isomorphism requirement
between the derivation trees may be too strong as well.  The former
problem admits of a straightforward solution, which we describe below.
The latter does not; we describe the symptoms of the problem but leave
its resolution as an open issue for further research.

\subsection{Multiple Adjunction}
\label{sec:multiple}

\begin{figure}
\qbox{\psfig{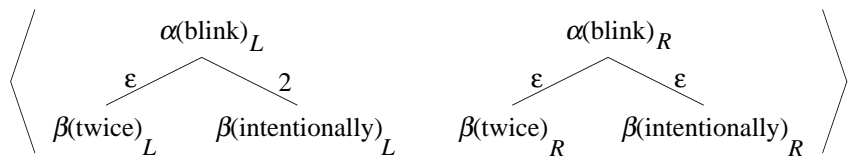}}
\caption{Derivation tree pair for the grammar of
Figure~\protect\ref{fig:blink-gram}.}
\label{fig:blink-map}
\end{figure}

Consider the synchronous TAG analysis of the semantics of adverbs
given in Figure~\ref{fig:blink-gram}.  This grammar is intended to
allow for the ambiguity of strings such as \used{John intentionally
blinked twice} as shown in Figure~\ref{fig:blink-derivs}. As
previously mentioned, the ambiguity arises from the ability to perform
two rewriting steps at the same node, the root $F$ node in the
elementary tree $\alpha(blink)_R$ corresponding to the word \word{blinked}.
Under the natural definition, however, this would entail a derivation
tree pair of the geometry given in Figure~\ref{fig:blink-map}.  But
the right derivation tree is ill-formed, as it violates the
prohibition against multiple adjunctions at a single address.

It was the desire to model semantic ambiguity through violations of
the prohibition that led us originally to a rewriting --- as opposed
to a derivation tree --- approach to defining synchronous TAG
derivation.  Thus, the deviation from the natural definition of
synchronous derivation was necessary because we required the ability
of two elementary trees to be adjoined at the same node.
Unfortunately, the rewriting interpretation of TAGs is a very
inelegant way in which to get this ability, leading as it does to the
problems described in Section~\ref{sec:problems}.  Nonetheless,
without this ability, the utility of synchronous TAGs is severely
diminished.

\begin{figure}
\qbox{\psfig{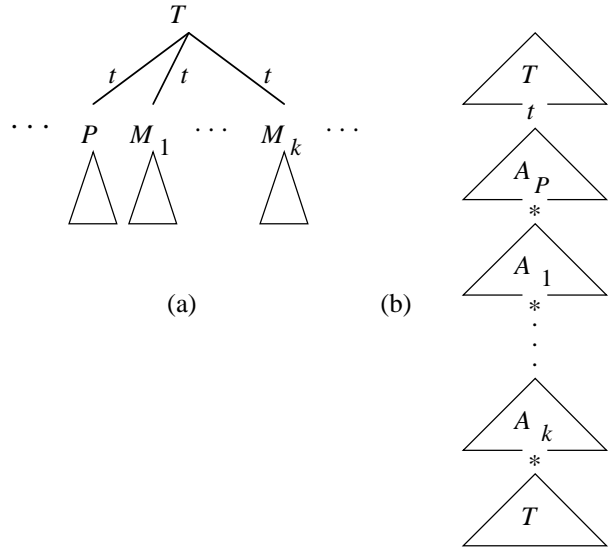}}
\caption{Interpretation of derivations with multiple adjunctions at a
single node.  In this case, several modifier trees $M_1$ through $M_k$
have been adjoined at node $t$ in tree $T$, along with a single
predicative node $P$.  The derived trees associated with $P$ and $M_1$
through $M_k$, namely $A_P$ and $A_1$ through $A_k$ appear in the
derived tree in that order.}
\label{fig:interp}
\end{figure}

For quite separate reasons, Schabes and I have been examining
alternatives to Vijay-Shanker's definition of TAG derivation so as to
allow for multiple adjunctions of certain auxiliary trees at the same
node.  Our solution \cite{full-alt-deriv} divides the class
of auxiliary trees into two types, {\em modifier} trees and {\em
predicative} trees, of which only the former allow such multiple
adjunctions.  In Vijay-Shanker's definition of derivation, a
derivation tree is well-formed if no two auxiliary trees are adjoined
at the same node in the same tree.  In our revised definition, a
derivation tree is well-formed if no two {\em predicative} auxiliary
trees are adjoined at the same node in the same tree.  Furthermore, so
as to determinately specify a derived tree, all modifier trees that
are adjoined at the same node in the same tree are ordered with
respect to one another.  Figure~\ref{fig:interp} shows the
interpretation, in terms of derived tree (\ref{fig:interp}b), of a
derivation tree (\ref{fig:interp}a) with multiple adjunctions at a
single node.  In essence, this diagram gives the interpretation of the
operation $I[A_1/t_1, \ldots, A_k/t_k]$ when the $t_k$ are not
disjoint.

The existence of the revised definition of derivation vitiates the
argument for the flat definition of synchronous TAG derivation.
Rather, a direct definition is now possible along the previous lines.
The only difference is that $D_L$ and $D_R$ are taken to be
well-formed derivation trees of the new variety.  Taking the trees
$\beta(twice)_R$ and $\beta(intentionally)_R$ to be modifier trees,
the synchronous derivation in Figure~\ref{fig:blink-map} is
well-formed.  The two possible orderings of the child nodes adjoining
at address $\epsilon$ provide for the two readings of the ambiguous
sentence.

\subsection{The Isomorphism Requirement}

A potentially more severe (and certainly more subtle) problem results
from the requirement of isomorphism between $D_L$ and $D_R$.  There
seem to be certain applications of synchronous TAGs for which this
requirement is too strong.  In this section, we present a taxonomy of
potential counterexamples to isomorphism, organized by the ``shape''
of the nonisomorphic part of the mapping between the derivation trees.
The examples are drawn from both technological application of
synchronous TAG to the problem of defining translations between
languages and application of synchronous TAG to the modeling of
natural language semantics.  It may turn out that different
applications provide different amounts of pressure to loosen the
isomorphism requirement in differing ways.  Although we discuss
several possible approaches to resolving this issue, we leave to
further work whether a satisfactory solution for a given application
can be found, and if so, what that solution might be.

\subsubsection*{Many-to-One Mappings}

\begin{center}
\mbox{\psfig{figure=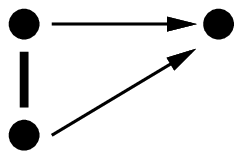}}
\end{center}

The simplest examples are cases in which an atomic construction in one
language is compound in another.  For example, \namecite{asj90} point
out that the English adverbial \word{hopefully} is translated by the French
phrase \word{on esp\`{e}re que}.  Whereas the English corresponds to a
single elementary tree, the French corresponds to a tree derived by
substituting the elementary tree for \word{on} as the NP argument of
\word{esp\`{e}re}.  Such examples argue for the ability to allow the
mapping between the left and right derivation trees to be relaxed from
a strict isomorphism.

One might think (as indeed the present author did before penetrating
discussions with Anthony Kroch) that a mismatch such as this shows
that the isomorphism requirement must also be too strong for the
purpose of modeling natural language semantics, for if these two
constructions --- \word{hopefully} and \word{on esp\`{e}re que} --- have the
same semantics, then at least one of the two (if not both) must
exhibit a mismatch between the natural language derivation and a
derivation of its logical form.  The error in this reasoning follows
from the assumption that the relationship of ``corresponds as an
appropriate translation'' (in the sense in which bilingual
dictionaries record such facts) is tantamount to ``means the same
as''.  This assumption is highly suspect.  Bilingual dictionaries do
not codify perfect translations in any sense, if such a notion is even
coherent.

However, mismatches of this variety may also be found in applications
to directly modeling natural-language semantics.  For instance, the
transduction relationship between a compound idiom (such as \used{kick
the bucket}) and its atomic semantics (given, e.g., by a simple
predication of $die$) might be thought to be of this form.

\subsubsection*{Elimination of Dominance}

\begin{center}
\mbox{\psfig{figure=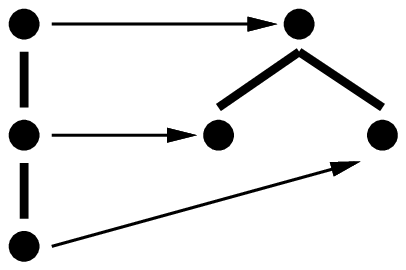}}
\end{center}

Even when the number of nodes in the paired derivation trees is the
same, they may exhibit different structure.  Nodes participating in a
domination relationship in one tree may be mapped to nodes neither of
which dominates the other.

Abeill\'e (personal communication) has noted a potential example of
such a mismatch.  For instance, in the sentence

\enumsentence{
Le docteur soigne les dents de Jean.\\
The doctor treats Jean's teeth.}

\noindent the subphrase \word{de Jean} is substituted into the
\word{dents} tree syntactically, and arguably modifies the semantics
of that tree as well.  However, the cliticized version of the sentence

\begin{figure}
\qbox{\psfig{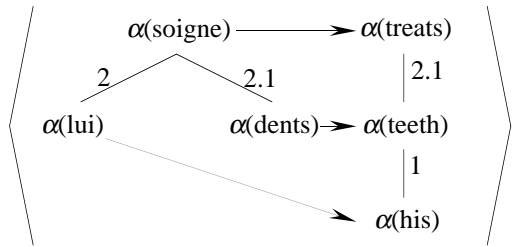}}
\caption{Schematic derivation tree pair for example
(\protect\ref{ex:clitic}).  The arrows show the required mapping
between the derivations, which is not an isomorphism.}
\label{fig:clitic-map}
\end{figure}

\enumsentence{\label{ex:clitic}
Le docteur lui soigne les dents.\\
The doctor treats his teeth.}

\noindent involves syntactic adjunction of the clitic \word{lui} in the
tree for \word{soigne}, although the translation into English, as before,
places the pronoun within the object NP of the verb.  Schematically,
the derivation trees should show the geometry given in
Figure~\ref{fig:clitic-map}.  Note that the separate derivations are
not isomorphic; a sibling relation in one tree corresponds to a
domination relation in the other.

Again, examples may be found in the arena of semantic interpretation.
Although the argumentation is much more complex, and well beyond the
scope of this paper, similar relationships arise in the context of
modeling quantifier scope ambiguity.

\subsubsection*{Inversion of Dominance}

\begin{center}
\mbox{\psfig{figure=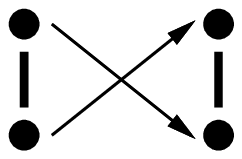}}
\end{center}

An even more extreme relationship, in which domination relationships
are not only introduced but actually inverted, is exemplified by the
French sentence and its English translation given in (\ref{ex:monte}),
and discussed by \namecite{whitelock-shake}.

\enumsentence{\label{ex:monte}
 Jean monte la rue en courant.\\
 John runs up the street.}

\begin{figure}
\qbox{\psfig{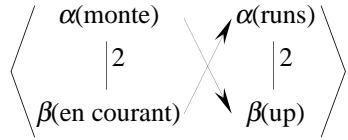}}
\caption{Schematic derivation tree pair for example
(\protect\ref{ex:monte}).  The arrows show the required mapping
between the derivations, which is not an isomorphism.}
\label{fig:courant-map}
\end{figure}

\noindent In this example, the phrase \word{en courant} adjoins as an
adverbial modifier to the verb \word{monte}.  Presumably, \word{en courant}
would be paired with the English \word{runs} and \word{monte} with the English
\word{up}.  But the derivation tree for the English sentence would not then
have the isomorphic structure in which \word{runs} adjoins or substitutes
into \word{up}, at least under the most natural analysis.  Rather, the
converse should hold; \word{up} should be inserted into \word{runs}.
Figure~\ref{fig:courant-map} shows the derivation tree pair
schematically, including the nonisomorphism mapping between the trees.

We know of no example of inversion of dominance in applications to
natural-language semantics.

\subsection{Relaxing Isomorphism}

In many of the above examples, although the mapping among derivation
nodes is not an isomorphism, the deviation from isomorphism is nicely
bounded, so that they could be well handled by allowing bounded
subderivations to be considered elementary for the purpose of defining
the relationship between the trees.  In using synchronous TAGs as a
model for language translation, that is, essentially to specify a
bilingual lexicon, it is not surprising that bounded subderivations in
one language are paired as a whole with bounded subderivations in
another.  Indeed, this is the modus operandi for traditional bilingual
dictionaries.  The Harper/Collins/Robert English-French dictionary
provides an entry for \word{to [run] down/in/off} with translation
\word{descendre/entrer/partir en courant} essentially providing the
mapping between the pertinent subderivations.  Similarly, the
pertinent entry under \word{hopefully} specifies the translation of
\word{[hopefully] it won't rain} as \word{on esp\`{e}re qu'il ne va pas
pleuvoir}, providing implicitly the subderivation mapping of
\word{hopefully} in its presentential position with \word{on esp\`{e}re
que}.  For the most part, placing the isomorphism at the level of
certain primitive and bounded subderivations is plausible,
sufficiently expressive,\footnote{The French clitic example, however,
remains problematic.  The relation between the clitic and the NP which
it is semantically related to seems to be potentially unbounded.} and
retains the advantages described in Section~\ref{sec:advantages}.

If further relaxation of the isomorphism requirement is to be allowed,
some method of controlling the relationship between the pair
derivations will be needed.  Owen Rambow and Giorgio Satta (personal
communication) have conjectured that an approach along the lines of
control grammars might be useful. This possibility, though
tantalizing, remains to be explored.

Whitelock's method of ``shake-and-bake'' translation
\cite{whitelock-shake}, under which translation involves reusing
the same components but under different relationships, seems to
correspond to a version of synchronous TAGs in which there is no
constraint on the geometries of the derivation trees, the only
requirement being that they are constructed from paired elements.
This extreme version of relaxing the isomorphism requirement may in
the end be necessary.

The exact nature of the relationship between paired derivation
trees must remain for future work.

\section{Conclusion}

We have introduced a simple, natural definition of synchronous
tree-adjoining derivation, based on isomorphisms between standard
tree-adjoining derivations, that avoids the expressivity and
implementability problems of the original rewriting definition.  The
decrease in expressivity, which would otherwise make the method
unusable, is offset by the incorporation of an alternative definition
of standard tree-adjoining derivation, previously proposed for
completely separate reasons, that allows for multiple adjunctions at a
single node in an elementary tree.  The increased flexibility from the
ability to perform such multiple adjunctions makes it conceivable to
entertain using the natural definition of synchronous derivation.
Nonetheless, some remaining problematic cases call for yet more
flexibility in the definition; the isomorphism requirement may have to
be relaxed.  It remains for future research to tune the exact
requirements on the allowable mappings.

\subsubsection*{Acknowledgements}

The research described in this paper was made possible in part by a
Presidential Young Investigator grant IRI-91-57996 from the National
Science Foundation and matching funds from Xerox Corporation.  An
early version of this paper was presented at the Second Workshop on
Tree-Adjoining Grammars, in Philadelphia, PA, in August 1992.  The
author is indebted to the following people for helpful discussions on
the subject matter of this paper: Anne Abeill\'{e}, Aravind Joshi,
Owen Rambow, Giorgio Satta, Yves Schabes, K. Vijay-Shanker, David
Weir, and Peter Whitelock.

{
\bibliographystyle{fullname}

}

\end{document}